\documentclass{article}

\usepackage{spconf,amsmath,amssymb,graphicx,multirow,longtable,array,tabularx,xcolor}

\DeclareMathOperator*{\argmax}{arg\,max}

\usepackage{comment}
\usepackage[ruled, lined, linesnumbered, commentsnumbered, longend]{algorithm2e}

\title{A MODEL-AGNOSTIC ACTIVE LEARNING APPROACH FOR \\ANIMAL DETECTION FROM CAMERA TRAPS}
\name{Thi Thu Thuy Nguyen and Duc Thanh Nguyen}
\address{Machine Intelligence Lab, School of Information Technology, Deakin University}

\begin{document}
%\ninept
%
\maketitle
\begin{abstract}
Smart data selection is becoming increasingly important in data-driven machine learning. Active learning offers a promising solution by allowing machine learning models to be effectively trained with optimal data including the most informative samples from large datasets. Wildlife data captured by camera traps are excessive in volume, requiring tremendous effort in data labelling and animal detection models training. Therefore, applying active learning to optimise the amount of labelled data would be a great aid in enabling automated wildlife monitoring and conservation. However, existing active learning techniques require that a machine learning model (i.e., an object detector) be fully accessible, limiting the applicability of the techniques. In this paper, we propose a model-agnostic active learning approach for detection of animals captured by camera traps. Our approach integrates uncertainty and diversity quantities of samples at both the object-based and image-based levels into the active learning sample selection process. We validate our approach in a benchmark animal dataset. Experimental results demonstrate that, using only 30\% of the training data selected by our approach, a state-of-the-art animal detector can achieve a performance of equal or greater than that with the use of the complete training dataset.
\end{abstract}
\section{Introduction}
\label{sec:intro}

Recent advances in computer vision and machine learning have created new capacities in ecological research. Automated wildlife detection, empowered by deep learning, has become crucial for conservation and biodiversity assessments~\cite{2018Norouzzadeh}. However, the performance of machine learning models heavily depends on the availability of labelled data, which are often difficult, time-consuming, and costly to obtain. This problem becomes more challenging in cases where the wildlife species are rare to capture from camera traps~\cite{Beery2021iWildCam, Angela_2025}. Consequently, there has been increasing interest in leveraging active learning (AL) to reduce the labelling burden while maintaining high performance of dowstream tasks, e.g., animal detection and recognition~\cite{Norouzzadeh2019ADA}.

AL refers to a set of machine learning techniques that aim to select an optimal dataset used to train a target machine learning model~\cite{DBLP:journals/corr/abs-2405-00334}. AL techniques can be categorised into model-specific and model-agnostic approaches. Model-specific methods require full access to the target model, e.g., accessing the target model's intermediate layers~\cite{DBLP:conf/cvpr/YangHC24}, or modifying the target model's architecture~\cite{ZHANG2024127883}. However, such a requirement is not always satisfied due to privacy-preserving concerns~\cite{DBLP:conf/eccv/YuanWLXL22}. In this paper, we focus on model-agnostic AL where only outputs from the target model are used for AL. 

For wildlife detection from camera traps, animal images captured by the same camera trap have the same background. We found this observation particularly helpful for applying AL to wildlife detection, as this can help find a compact yet diverse set of images to effectively train an animal detector. As argued in~\cite{DBLP:conf/cvpr/GhiasiCSQLCLZ21}, the background information is crucial for the performance of object detection and instance segmentation methods. The authors also showed that one can augment the training data of a detector by simply cutting and pasting object images at different locations in the same background, reducing the number of training images. To exploit this property, we propose to integrate the box-level uncertainty of animal detections and the image-level diversity of camera trap images into the sampling process in AL. In particular, we make the following contributions in our work.
\begin{itemize}
    \item We propose a model-agnostic AL approach integrating both object-level uncertainty and image-level diversity for animal detection from camera traps. The object-level uncertainty of animal detections is outputted by a target animal detector in a model-agnostic manner (i.e., only outputs of the model are used while the model is preserved from being accessed). The image-level diversity of camera trap-captured images is measured via the distances between the images' embeddings produced by a task-agnostic pre-trained image encoder. 
    \item We devise two AL methods, assembling the uncertainty and diversity quantities in different ways.
    \item We conducted extensive experiments to validate our proposed AL methods and compared them with existing model-agnostic AL baselines in a benchmark camera trap wildlife dataset. 
\end{itemize}

\section{Related Work}
\label{sec:related_work}

AL methods select samples based on two common criteria: uncertainty and diversity. The uncertainty of a sample is often derived from the confidence score of the target model with respect to that sample. For instance, in~\cite{DBLP:conf/cvpr/YangHC24}, the uncertainty of a detection was calculated from the objectness score of that detection and the likelihood of the object class recognised in that detection. Brust et al.~\cite{DBLP:conf/visapp/BrustKD19} calculated the uncertainty from the difference between the probabilities for the best and second-best recognised object classes. Roy et al.~\cite{roy2018deep} proposed using the Shannon entropy of the detected object classes for the uncertainty. In~\cite{Zhang2024}, the uncertainty of a detection was estimated as the uncertainty of the model in predicting the size and location of that detection~\cite{Zhang2024}. In~\cite{ZHANG2024127883}, the variation of the target model's outputs to small changes made in samples was used for the uncertainty calculation.

While the uncertainty measurement is achieved from individual samples, diversity measurement is calculated with respect to a set of samples. For instance, the authors in~\cite{DBLP:conf/cvpr/WuC022, DBLP:conf/cvpr/YangHC24} encoded each detection with an embedding extracted from an intermediate layer of the target model. The diversity of detections was then measured via the cosine distance between their embeddings. However, these methods are computationally expensive, as they require calculating the distances between pairs of detections in each image and across images.

Existing AL approaches implement uncertainty and diversity quantities in different ways. Model-specific AL methods suppose that the target model is fully accessible. They make use of the internal information of the target model, e.g., intermediate layers~\cite{DBLP:conf/cvpr/YangHC24, DBLP:conf/cvpr/WuC022}, or modify the model architecture~\cite{ZHANG2024127883} during the AL process. However, access to the target model is not always feasible. As shown in~\cite{DBLP:conf/eccv/YuanWLXL22}, data leakage may occur from a pre-trained model. In addition, the target model may be preserved for commercial purposes. Model-agnostic methods, on the other hand, are independent of the target model, since they are inputted by outputs of the target model~\cite{roy2018deep, DBLP:conf/visapp/BrustKD19, DBLP:conf/colt/GajjarTXHML24}. However, without accessing to the target model, these model-agnostic methods can select samples based only on the uncertainty of the samples. In a camera trap-based animal detection application, background images captured by a camera trap are mostly identical (or similar). This makes the captured data redundant. An appropriate use of the background's diversity would be helpful to reduce the amount of camera trap data, saving the data labelling cost while improving the adaptivity of the target model to various background data.    

\section{Methods}
\label{sec:pagestyle}

\subsection{Overview}

Our problem of interest can be stated as follows. Given a target animal detector $f$ (e.g., a neural network) that aims to detect $C$ animal classes from an input image, and a set of unlabelled animal images $\mathcal{X}$, our objective is to select a subset $\hat{\mathcal{X}}$ of $B$ images from $\mathcal{X}$ to train the detector $f$. The above data sampling process can be repeated until a desirable performance (e.g., the mean average precision of the target detector on a validation set) is reached. The increment of samples $B$ (as known as the budget) is fixed at every iteration.

We focus on model-agnostic AL for animal detection, i.e., the target detector $f$ is not fully accessible during the data sampling process. Specifically, given an image $\mathbf{x} \in \mathcal{X}$, we assume that the outputs of the detector $f$ on $\mathbf{x}$, denoted as $\mathcal{D}^f(\mathbf{x})$, are the only available information. Following the conventional object detection paradigm, we define $\mathcal{D}^f(\mathbf{x}):=\{\mathbf{d}_i(\mathbf{x})=(\mathbf{b}_i(\mathbf{x}),s_i(\mathbf{x}),c_i(\mathbf{x}))\}$ where $\mathbf{d}_i$ is a detection described by a bounding box $\mathbf{b}_i \in \mathbb{R}^4$, a detection score $s_i \in \mathbb{R}$, and an animal class $c_i \in \{1,...,C\}$ that most likely appears in $\mathbf{b}_i$. The score $s_i$ can be calculated as the probability $p(c_i)$ that the class $c_i$ is recognised in $\mathbf{b}_i$~\cite{DBLP:journals/pami/RenHG017}, or as the product of the confidence score of $f$ with respect to $\mathbf{b}_i$ and $p(c_i)$~\cite{DBLP:conf/cvpr/RedmonF17}. Several AL methods, e.g.,~\cite{DBLP:conf/visapp/BrustKD19, DBLP:conf/cvpr/YangHC24}, require the class probability distribution of all object classes extracted from the last softmax layer of $f$ for each detection $\mathbf{d}_i$, in their inputs.

In our work, the criteria used for the image sampling in AL are the uncertainty of the target detector $f$ with respect to its detections in each image $\mathbf{x}$ and the diversity of $\mathbf{x}$. We denote $u(\mathbf{x})$ as the uncertainty of an image $\mathbf{x}$, and estimate it through the detection scores of all detections in $\mathbf{x}$. Formally, we define,
\begin{align}
    \label{eq:uncertainty}
    u(\mathbf{x}) = \frac{1}{|\mathcal{D}^f(\mathbf{x})|}\sum_{i=1}^{|\mathcal{D}^f(\mathbf{x})|} (1 - s_i(\mathbf{x}))
\end{align}
where $|\mathcal{D}^f(\mathbf{x})|$ is the cardinality of the detection set $\mathcal{D}^f(\mathbf{x})$.

Existing research measures the diversity of detections through the diversity of their representations extracted from an intermediate layer of the target detector~\cite{DBLP:conf/cvpr/WuC022, DBLP:conf/cvpr/YangHC24}. However, this approach is not applicable to the model-agnostic AL setting. Instead, we take advantage of a task-agnostic pre-trained image encoder to extracting image representations. Furthermore, we estimate the diversity at the image level since we observed that, the background takes a majority of a camera trap image while images captured by the same camera trap share the same background information. Image-level diversity thus would help identify a minimal yet discriminative set of backgrounds to effectively train the detector. Specifically, let $g$ be a task-agnostic image encoder and $\mathcal{X}'$ be an image set. We define the diversity of an image $\mathbf{x}$ to the set $\mathcal{X}'$ as,
\begin{align}
    \label{eq:diversity}
    v(\mathbf{x},\mathcal{X}')=\frac{1}{|\mathcal{X}'|}\sum_{\mathbf{x}' \in \mathcal{X}'} \delta(g(\mathbf{x}),g(\mathbf{x}'))
\end{align}
where $g(\mathbf{x})$ is an embedding of $\mathbf{x}$ generated by $g$, and $\delta$ is the Euclidean distance between two embeddings. In the following, we present two AL methods using the uncertainty and diversity quantities defined above.

\begin{algorithm}[ht]
    \SetKwFunction{isOddNumber}{isOddNumber}
    
    \SetKwInOut{KwIn}{Input}
    \SetKwInOut{KwOut}{Output}

    \KwIn{Data ($\mathcal{X}$), budget ($B$), number of iterations ($N$), target detector ($f$), image encoder ($g$)}
    \KwOut{Sampled image set ($\hat{\mathcal{X}}$)}

    $\hat{\mathcal{X}}$: randomly sampled from $\mathcal{X}$ and labelled by an annotator.

    \For{$i:=1$ \KwTo $N$}
    {
        $f \leftarrow$ train using $\hat{\mathcal{X}}$
    
        $\mathcal{U}:=\emptyset$\tcp*[f]{Uncertainty score set}
        
        $\mathcal{G}:=\emptyset$\tcp*[f]{Image embedding set}
        
        \For{$\mathbf{x} \in \mathcal{X} \backslash \hat{\mathcal{X}}$}{
            $\mathcal{U}:=\mathcal{U} \cup \{u(\mathbf{x})\}$

            $\mathcal{G}:=\mathcal{G} \cup \{g(\mathbf{x})\}$
        }

        $\{\mathcal{X}_1, ...,\mathcal{X}_B\}:=\text{K-means}(\mathcal{Y}=\mathcal{X}\backslash\hat{\mathcal{X}},\mathcal{E}=\mathcal{G},K=B)$

        \For{$k:=1$ \KwTo $B$}
        {
            $\mathbf{x}_k:=\argmax_{\mathbf{x} \in \mathcal{X}_k} u(\mathbf{x})$
        
            $\hat{\mathcal{X}}:=\hat{\mathcal{X}} \cup \{\mathbf{x}_k\}$

            $\mathcal{X}:=\mathcal{X}\backslash\{\mathbf{x}_k\}$
        }
    }

    \KwRet{$\hat{\mathcal{X}}$}
    
    \caption{\small We use an auxiliary procedure: K-means$(\mathcal{Y},\mathcal{E},K)$, taking three inputs: an image set $\mathcal{Y}$, a corresponding embedding set $\mathcal{E}$, and the number of clusters $K$. It then clusters all the embeddings into $K$ groups and classifies all the images into corresponding groups.}
    \label{alg:method_1}
\end{algorithm}

\subsection{Method 1: Diversity-driven uncertainty-based AL}

The core idea of this method is that, instead of performing data sampling on the entire dataset $\mathcal{X}$, we apply it only to key areas in $\mathcal{X}$. This can be realised in two steps: finding the most distinctive partition of $\mathcal{X}$, then applying uncertainty-based data sampling to the partitioned areas. Specifically, suppose that $\mathcal{P}(\mathcal{X})$ is a partition of $\mathcal{X}$, i.e., $\mathcal{P}(\mathcal{X})=\{\mathcal{X}_1,...,\mathcal{X}_K\}$ where $\forall k,j\in \{1,...,K\}$, $\mathcal{X}_k \subset \mathcal{X}$, $\bigcup \mathcal{X}_k=\mathcal{X}$, and $\mathcal{X}_k \cap \mathcal{X}_j = \emptyset$. The most distinctive partition of $\mathcal{X}$, denoted as $\mathcal{P}^*(\mathcal{X})$, can be determined based on the diversity of all the images in the subsets of $\mathcal{X}$ with respect to each other subset. Mathematically, we define, 
\begin{align}
    \label{eq:partition}
    \mathcal{P}^*(\mathcal{X}) = \argmax_{\mathcal{P}(\mathcal{X}) \in \mathrm{P}(\mathcal{X})} \sum_{\mathcal{X}_k \in \mathcal{P}(\mathcal{X})} \sum_{\mathbf{x} \in \mathcal{X}_k} v(\mathbf{x},\mathcal{X}_k)
\end{align}
where $\mathbf{P}(X)$ represents the space of all possible partitions of $\mathcal{X}$ and $v(\mathbf{x}_j,\mathcal{X}_k)$, as defined in Eq.~(\ref{eq:diversity}), represents the diversity of the image $\mathbf{x}_j$ to the image set $\mathcal{X}_k$.

Solving the problem in Eq.~(\ref{eq:partition}) using a brute-force search has high complexity, as the number of partitions to be investigated is a Bell number~\cite{Halmos_1974}. Here, we solve Eq.~(\ref{eq:partition}) using a relaxed fashion (see Algorithm~\ref{alg:method_1}). In particular, we limit the search space $\mathbf{P}(X)$ to partitions with $B$ subsets. Moreover, instead of maximising the inter-set distances represented by the diversity $v$, we minimise intra-set distances. This relaxation can be achieved by applying the $K$-means algorithm (with $K=B$) to divide $\mathcal{X}$ into $B$ clusters. After clustering, the representative image for each cluster $\mathcal{X}_k$ is denoted as $\mathbf{x}_k$ and selected as the one with the highest uncertainty score within the cluster, i.e., 
\begin{align}
    \label{eq:sampling_method_1}
    \mathbf{x}_k = \argmax_{\mathbf{x} \in \mathcal{X}_k} u(\mathbf{x})
\end{align}

\subsection{Method 2: Uncertainty-driven diversity-based AL}

This method samples an image by maximising the image's uncertainty with respect to the target detector $f$ and, at the same time, maximising the image's diversity to already-sampled
images (i.e., images sampled in previous iterations). The crucial factor of this method is the sampling score that can integrate both the uncertainty and diversity quantities. Suppose that $\hat{\mathcal{X}}$ is the current sampled set and $|\hat{\mathcal{X}}|<B$ (i.e., there is still room for more images to be selected), we define the sampling score $z(\mathbf{x})$ for each unlabelled image $\mathbf{x} \in \mathcal{X} \backslash \hat{\mathcal{X}}$ as follows,
\begin{align}
    \label{eq:sampling_method_2}
    z(\mathbf{x}|\hat{\mathcal{X}})=(1-\alpha)u(\mathbf{x})+\alpha v(\mathbf{x},\hat{\mathcal{X}})
\end{align}
where $\alpha \in [0,1]$ is a parameter. 

We perform the sampling in a
greedy fashion. In particular, we start each iteration with the image
with the highest uncertainty score. We then continue to select the following images one by one
based on their sampling scores to the already-labelled images (see Algorithm~\ref{alg:method_2}). 

As shown in Eq.~(\ref{eq:sampling_method_2}), $\alpha$ plays a role as a weight factor to balance the uncertainty and diversity terms. In this paper, instead of fixing $\alpha$, we tune it adaptively with the remaining unlabelled data in every iteration. Specifically, suppose that the initial value of $\alpha$ in the first iteration (iteration $0$) is $\alpha^{(0)}$, we then update $\alpha^{(i)}$ in following iterations $i$ ($i > 0$) using the rule below,
\begin{align}
    \label{eq:alpha}
    \alpha^{(i)} \leftarrow \alpha^{(i-1)} - \frac{B}{2|\mathcal{X}\backslash\hat{\mathcal{X}}|}
\end{align}
where $|\mathcal{X}\backslash\hat{\mathcal{X}}|$ is the number of unlablled images.

We initialise $\alpha^{(0)}=0.5$, a neutral value indicating that both the uncertainty and diversity terms contribute equally to the sampling score $z(\mathbf{x})$. However, the values of $\alpha$ in the following rounds are updated accordingly. The rationale for the update rule is that, when more images are labelled, the diversity of the unlabelled set (i.e., $\mathcal{X}\backslash\hat{\mathcal{X}}$) naturally decreases due to a lower data density. In addition, since the target detector gets improved after a number of training iterations, its uncertainty score estimation becomes more reliable, and samples with high uncertainty scores are thus more critical to the detection system. We found that this simple update rule works effectively in practice.

\begin{algorithm}
    \SetKwFunction{isOddNumber}{isOddNumber}
    \SetKwInOut{KwIn}{Input}
    \SetKwInOut{KwOut}{Output}

    \KwIn{Data ($\mathcal{X}$), budget ($B$), number of iterations ($N$), target detector ($f$), image encoder ($g$)}
    \KwOut{Sampled image set ($\hat{\mathcal{X}}$)}

    $\hat{\mathcal{X}}$: randomly sampled from $\mathcal{X}$ and labelled an annotator

    \For{$i:=1$ \KwTo $N$}
    {
        $f \leftarrow$ train using $\hat{\mathcal{X}}$
    
        $\mathcal{U}:=\emptyset$\tcp*[f]{Uncertainty score set}
        
        $\mathcal{G}:=\emptyset$\tcp*[f]{Image embedding set}
        
        \For{$\mathbf{x} \in \mathcal{X} \backslash \hat{\mathcal{X}}$}{
            $\mathcal{U}:=\mathcal{U} \cup \{u(\mathbf{x})\}$

            $\mathcal{G}:=\mathcal{G} \cup \{g(\mathbf{x})\}$
        }

        $\mathbf{x}_1:=\argmax_{\mathbf{x} \in \mathcal{X}\backslash\hat{\mathcal{X}}} u(\mathbf{x})$\tcp*[f]{1st sample}

        $\hat{\mathcal{X}}:=\hat{\mathcal{X}} \cup \{\mathbf{x}_1\}$

        \For{$k:=2$ \KwTo $B$}
        {
            $\mathbf{x}_k:=\argmax_{\mathbf{x} \in \mathcal{X} \backslash \hat{\mathcal{X}}} z(\mathbf{x}|\hat{\mathcal{X}})$
        
            $\hat{\mathcal{X}}:=\hat{\mathcal{X}} \cup \{\mathbf{x}_k\}$

            $\mathcal{X}:=\mathcal{X}\backslash\{\mathbf{x}_k\}$
        }
    }

    \KwRet{$\hat{\mathcal{X}}$}
    
    \caption{\small We perform the sampling process in AL using sampling scores in a greedy fashion.}
    \label{alg:method_2}
\end{algorithm}

\section{Experiments}

\subsection{Dataset}

\begin{figure}[!ht]
    \centering
    \includegraphics[width=0.42\linewidth]{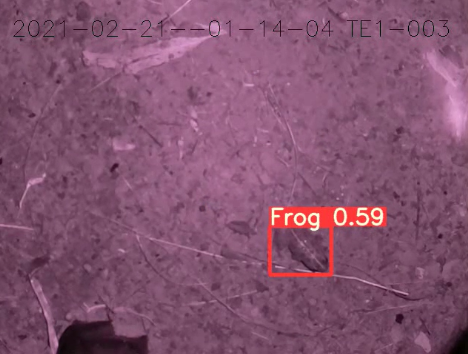}
    \includegraphics[width=0.43\linewidth]{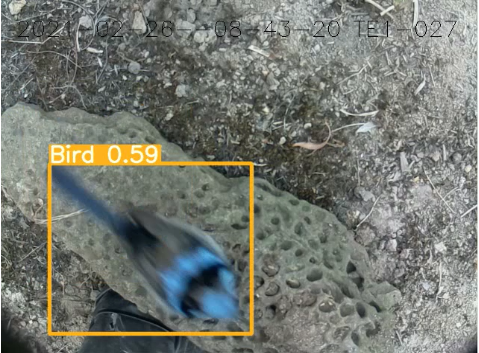}
    \includegraphics[width=0.43\linewidth]{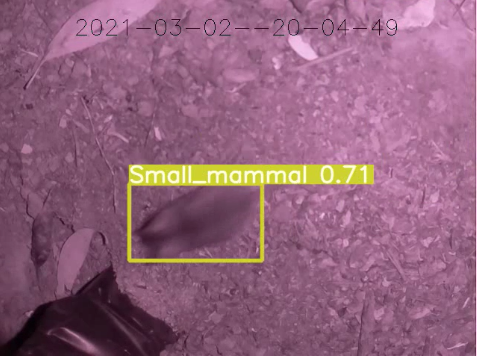}
    \includegraphics[width=0.43\linewidth]{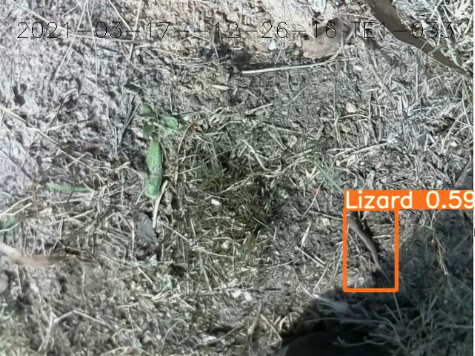}
    \includegraphics[width=0.43\linewidth]{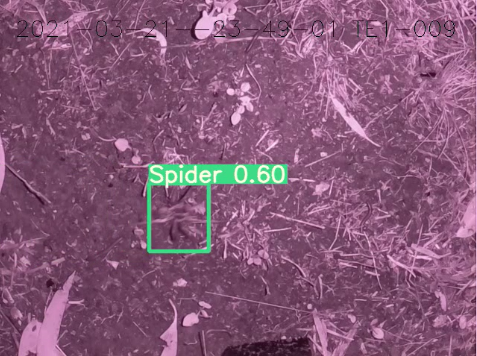}
    \includegraphics[width=0.43\linewidth]{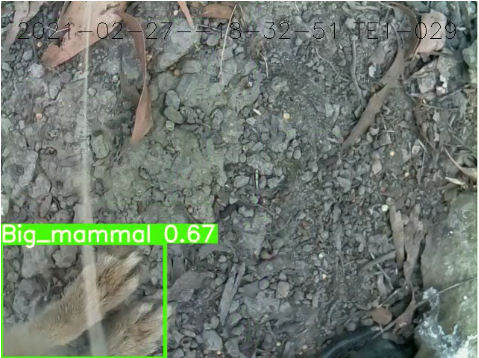}
    \includegraphics[width=0.43\linewidth]{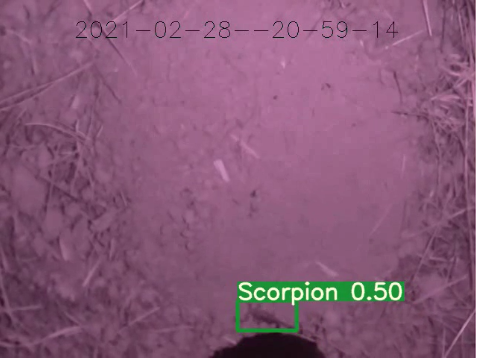}
    \caption{Example images from the SAWIT dataset~\cite{Nguyen2023SAWITAS} with detection results by the YOLO detector~\cite{DBLP:conf/cvpr/RedmonF17}.}
    \label{fig:SAWIT}
\end{figure}

We validated our method on SAWIT~\cite{Nguyen2023SAWITAS}, a benchmark dataset of small-sized animals captured from camera traps in the wild. This dataset contains 34,434 images across seven distinct categories, including frog, lizard, bird, small mammal, big mammal (medium-sized mammal), spider, and scorpion. The dataset also includes various challenging scenarios, such as occlusions, blurriness, and cases where the animals are camouflaged within dense vegetation. We followed the train/test split used in~\cite{Nguyen2023SAWITAS}, where 24,344 images were used for training and the remaining 11,000 images were used for validation. We present some example images of the SAWIT dataset in Fig.~\ref{fig:SAWIT}. As shown, the animals take small portions in the camera trap images. In addition, the images captured by the same camera have almost identical background, making the data redundant in terms of diversity.

\subsection{Experimental Setup}

We tested our method with YOLOv8~\cite{yolov8_ultralytics}, the 8-th version of the YOLO detector~\cite{DBLP:conf/cvpr/RedmonF17}. YOLOv8 is well-known for its robustness in terms of detection accuracy and processing speed, proven in different datasets and applications. We used YOLOv8 in a black-box manner, i.e., only the outputs (bounding boxes) returned by the detector are used. For the agnostic-task image encoder, we adopted the model of VGG-16 pre-trained on ImageNet\footnote{https://pytorch.org/vision/main/models/generated/torchvision.models.vgg16.html}. This model encodes an input image into a 4,096-dimensional vector. 

For the AL setting, we initialised our target detector (i.e., YOLOv8) with a set of 2,434 images (10\% of the entire training data). This initial set was randomly sampled but guaranteed to include all the animal classes. The AL process was then iteratively applied, with a budget of 1,712 images (5\% of the entire training data) for each iteration. This process continued until half of the training set had been selected. 

\begin{table*}[ht]
\centering

\resizebox{\textwidth}{!}{ % Resize table to fit within one column width
\begin{tabular}{|c|c|c|c|c|c|c|c|c|c|c|c|}
\hline
\textbf{Iteration} & \textbf{Ratio} & \textbf{Random} & \textbf{Uncertainty} &\textbf{\cite{roy2018deep}-min} & \textbf{\cite{roy2018deep}-max} & \textbf{\cite{roy2018deep}-sum} & \textbf{\cite{DBLP:conf/visapp/BrustKD19}-sum} & \textbf{\cite{DBLP:conf/visapp/BrustKD19}-avg} & \textbf{\cite{DBLP:conf/visapp/BrustKD19}-max} &  \textbf{Method 1} & \textbf{Method 2} \\
\hline
0  & 10\%  & 0.532 & 0.532 & 0.532 & 0.532 & 0.532 & 0.532 & 0.532 & 0.532 & 0.532 & 0.532 \\
\hline
1  & 15\%  & 0.532 & 0.575 & 0.575 & 0.570  & \textbf{0.588} & 0.587 & 0.566 & 0.564 & \underline{0.586} & 0.580  \\
\hline
2  & 20\%  & 0.579 & 0.580  & 0.588 & 0.591 & 0.593 & 0.581 & 0.561 & 0.563 & \textbf{0.605} & \underline{0.598} \\
\hline
3  & 25\%  & \underline{0.609} & 0.566 & 0.560  & 0.568 & 0.601 & 0.583 & 0.593 & \textbf{0.611} & 0.587 & 0.588 \\
\hline
4  & 30\%  & 0.591 & 0.585 & 0.585 & 0.578 & 0.580  & 0.585 & \underline{0.596} & 0.569 & \textbf{0.630}
  & 0.576 \\
\hline
5  & 35\%  & 0.593 & \underline{0.610}  & 0.594 & 0.603 & 0.606 & 0.607 & 0.582 & 0.601 & 0.593 & \textbf{0.612} \\
\hline
6  & 40\%  & 0.565 & \underline{0.620}  & 0.585 & 0.598 & 0.584 & 0.577 & 0.584 & 0.612 & 0.580  & \textbf{0.624} \\
\hline
\end{tabular}
} % End resizebox

\text{mAP@50}

\resizebox{\textwidth}{!}{
\begin{tabular}{|c|c|c|c|c|c|c|c|c|c|c|c|}
\hline
\textbf{Iteration} & \textbf{Ratio} & \textbf{Random} & \textbf{Uncertainty} &\textbf{\cite{roy2018deep}-min} & \textbf{\cite{roy2018deep}-max} & \textbf{\cite{roy2018deep}-sum} & \textbf{\cite{DBLP:conf/visapp/BrustKD19}-sum} & \textbf{\cite{DBLP:conf/visapp/BrustKD19}-avg} & \textbf{\cite{DBLP:conf/visapp/BrustKD19}-max} &  \textbf{Method 1} & \textbf{Method 2} \\
\hline
0  & 10\%  & 0.320  & 0.320  & 0.320  & 0.320  & 0.320  & 0.320  & 0.320  & 0.320  & 0.320  & 0.320 \\
\hline
1  & 15\%  & 0.304 & 0.347 & 0.347 & 0.345 & 0.347 & 0.340  & 0.327 & 0.338 & \textbf{0.363} & \underline{0.347} \\
\hline
2  & 20\%  & 0.343 & 0.349 & 0.353 & 0.364 & 0.355 & 0.343 & 0.328 & 0.341 & \textbf{0.370}  & \underline{0.356} \\
\hline
3  & 25\%  & 0.357 & 0.346 & 0.330  & 0.338 & 0.352 & 0.357 & 0.352 & \underline{0.359} & \textbf{0.366} & \underline{0.359} \\
\hline
4  & 30\%  & 0.349 & \underline{0.355} & 0.352 & 0.344 & 0.350  & 0.353 & \underline{0.355} & 0.347 & \textbf{0.386} & 0.347 \\
\hline
5  & 35\%  & 0.358 & 0.364 & 0.354 & 0.371 & 0.362 & 0.367 & 0.350  & \underline{0.368} & 0.358 & \textbf{0.370}  \\
\hline
6  & 40\%  & 0.344 & \textbf{0.372} & 0.356 & 0.367 & 0.347 & 0.351 & 0.353 & 0.369 & \underline{0.370}  & 0.369 \\
\hline
\end{tabular}
} % End resizebox

\text{mAP@50-95}

\caption{mAP@50 and mAP@50-95 of the target detector (YOLOv8) trained with the data selected by our methods and by other baselines. For each iteration (except iteration $0$), top-1 and top-2 performances are highlighted in bold and underline.}

\label{tab:results}
\end{table*}

\subsection{Baselines}

We compared our method with the random sampling approach and existing model-agnostic AL baselines. For the random sampling approach, 1,712 images were randomly selected at each iteration. For the other model-agnostic AL baselines, we compared our work with the simply use of uncertainty scores. We also re-implemented the methods in~\cite{roy2018deep} and in~\cite{DBLP:conf/visapp/BrustKD19}. The method in~\cite{roy2018deep} uses the entropy of detected object classes to create sampling criteria, resulting in three sampling strategies: minimum-maximum entropy selection, maximum entropy selection, and sum entropy selection. The method in~\cite{DBLP:conf/visapp/BrustKD19} follows the uncertainty-based sampling approach, in which the uncertainty of a detection is measured as the difference between the probabilities for the best and second-best recognised object classes in that detection. This method also implements three different ways to calculate the uncertainty score of an image from the uncertainty scores of detections within the image, including summing, averaging, and maximising the detection scores. 

\subsection{Results}

We evaluated our proposed methods and existing baselines through the performance of the target detector trained with the data selected by our methods and existing ones. We also evaluated these AL methods for 6 iterations (excluding the initial one) with a fixed budget (5\% of the entire training data per iteration), resulting in proportions of the selected data from 15\% to 40\%. We measured the mean-average precision (mAP) of the target detector (after training) at two commonly used intersection-over-union (IoU) thresholds: 50\% (denoted as mAP@50) and [50\%,95\%] (denoted as mAP@50-95). 

We report the performance of the target detector in the above evaluations in Table~1. As shown in the results, our methods 1 and 2 often interchangeably take the 1-st and 2-nd rank. Method 1 also shows its advantage compared with the sole use of uncertainty scores (column 4). This is evident for the usefulness of the diversity-driven sampling approach where the partition of the sampled space helps lead the uncertainty-based sampling to more distinctive samples. Compared with method 1, method 2 shows more prominence in the later iterations (iterations 5 and 6). In general, method 1 achieves an overall better performance among all other methods. We also found an interesting point, that is, the target model achieves 0.611 for mAP@50 and 0.371 for mAP@50-90 when trained on the entire dataset. Meanwhile, the target detector obtains 0.630 for mAP@50 and 0.386 for mAP@50-90, when trained with only 30\% data selected by method 1, which exceeds the performance of the full training. This means that, by applying AL, one may need to train their detector with a small portion of the training data while achieving their highest performance.

All experiments were carried on 2 NVIDIA GeForce RTX2080Ti GPUs. The training time to train the target detector varied from 1 to 1.5 hours, with respective to different numbers of training data added per iteration (from 10\% to 40\%). Embedding extraction took around 15 mins on the entire training dataset. Embedding clustering with K-means algorithm took around 20 mins. Method 2 requires calculations of the distances between embeddings (see Eq.~(\ref{eq:diversity})). However such calculations are done only once, and took around 20 mins. In all, both methods 1 and 2 completed one data sampling iteration in around 30 mins. 

\section{Conclusion}
\label{sec:majhead}

The key achievement of our AL approach is its ability to maximise the performance of an animal detector with a significantly reduced amount of labelled data through a simple and black-box setting. This is enabled by incorporating both uncertainty and diversity quantities in the sampling process. As shown in experimental results, a target detector can achieve the highest accuracy (even exceeding the accuracy of the full training) when trained with only 30\% data. Our approach is not without limitations. Firstly, missed detections are not taken into account. Secondly, the approach does not consider the class balancing of the sampled images during the AL.  

\bibliographystyle{IEEEbib}
\bibliography{strings}

\begin{thebibliography}{10}

\bibitem{2018Norouzzadeh}
Mohammad~Sadegh Norouzzadeh, Anh Nguyen, Margaret Kosmala, Alexandra Swanson, Meredith~S. Palmer, Craig Packer, and Jeff Clune,
\newblock ``Automatically identifying, counting, and describing wild animals in camera-trap images with deep learning,''
\newblock {\em Proceedings of the National Academy of Sciences}, vol. 115, no. 25, pp. 5716--5725, 2018.

\bibitem{Beery2021iWildCam}
Sara Beery, Arushi Agarwal, Elijah Cole, and Vighnesh Birodkar,
\newblock ``The {iWildCam} 2021 competition dataset,''
\newblock {\em CoRR}, vol. abs/2105.03494, 2021.

\bibitem{Angela_2025}
Angela J.~L Pestell, Anthony~R Rendall, Robin~D. Sinclair, Euan~G. Rictchie, Duc~Thanh Nguyen, Dean~M. Corva, Anne~C. Eichholtzer, Abbas~Z. Kouzani, and Don~A. Driscoll,
\newblock ``Smart camera traps and computer vision improve detections of small fauna,''
\newblock {\em Ecosphere}, 2025.

\bibitem{Norouzzadeh2019ADA}
Mohammad~Sadegh Norouzzadeh, Dan Morris, Sara Beery, Neel Joshi, Nebojsa Jojic, and Jeff Clune,
\newblock ``A deep active learning system for species identification and counting in camera trap images,''
\newblock {\em Methods in Ecology and Evolution}, vol. 12, pp. 150--161, 2019.

\bibitem{DBLP:journals/corr/abs-2405-00334}
Dongyuan Li, Zhen Wang, Yankai Chen, Renhe Jiang, Weiping Ding, and Manabu Okumura,
\newblock ``A survey on deep active learning: Recent advances and new frontiers,''
\newblock {\em {IEEE} Transactions on Neural Networks and Learning Systems}, pp. 1--21, 2024.

\bibitem{DBLP:conf/cvpr/YangHC24}
Chenhongyi Yang, Lichao Huang, and Elliot~J. Crowley,
\newblock ``Plug and play active learning for object detection,''
\newblock in {\em {IEEE/CVF} Conference on Computer Vision and Pattern Recognition}, 2024, pp. 17784--17793.

\bibitem{ZHANG2024127883}
Licheng Zhang, Siew-Kei Lam, Dingsheng Luo, and Xihong Wu,
\newblock ``Employing feature mixture for active learning of object detection,''
\newblock {\em Neurocomputing}, vol. 594, pp. 127883, 2024.

\bibitem{DBLP:conf/eccv/YuanWLXL22}
Zhuowen Yuan, Fan Wu, Yunhui Long, Chaowei Xiao, and Bo~Li,
\newblock ``{SecretGen}: Privacy recovery on pre-trained models via distribution discrimination,''
\newblock in {\em European Conference on Computer Vision}, 2022, pp. 139--155.

\bibitem{DBLP:conf/cvpr/GhiasiCSQLCLZ21}
Golnaz Ghiasi, Yin Cui, Aravind Srinivas, Rui Qian, Tsung{-}Yi Lin, Ekin~D. Cubuk, Quoc~V. Le, and Barret Zoph,
\newblock ``Simple copy-paste is a strong data augmentation method for instance segmentation,''
\newblock in {\em {IEEE/CVF} Conference on Computer Vision and Pattern Recognition}, 2021, pp. 2918--2928.

\bibitem{DBLP:conf/visapp/BrustKD19}
Clemens{-}Alexander Brust, Christoph K{\"{a}}ding, and Joachim Denzler,
\newblock ``Active learning for deep object detection,''
\newblock in {\em International Joint Conference on Computer Vision, Imaging and Computer Graphics Theory and Applications}, 2019, pp. 181--190.

\bibitem{roy2018deep}
Soumya Roy, Asim Unmesh, and Vinay~P Namboodiri,
\newblock ``Deep active learning for object detection.,''
\newblock in {\em British Machine Vision Conference}, 2018, vol. 362, p.~91.

\bibitem{Zhang2024}
Zhipeng Zhang, Wenting Ma, Xiaohang Yuan, Yuan Hao, Meng Guo, Hongyi Tang, Zhiheng Zhou, and Zhenjie Yao,
\newblock ``Instance-aware uncertainty for active learning in object detection,''
\newblock in {\em IEEE International Conference on Image Processing}, 2024, pp. 298--304.

\bibitem{DBLP:conf/cvpr/WuC022}
Jiaxi Wu, Jiaxin Chen, and Di~Huang,
\newblock ``Entropy-based active learning for object detection with progressive diversity constraint,''
\newblock in {\em {IEEE/CVF} Conference on Computer Vision and Pattern Recognition}, 2022, pp. 9387--9396.

\bibitem{DBLP:conf/colt/GajjarTXHML24}
Aarshvi Gajjar, Wai~Ming Tai, Xingyu Xu, Chinmay Hegde, Christopher Musco, and Yi~Li,
\newblock ``Agnostic active learning of single index models with linear sample complexity,''
\newblock in {\em Annual Conference on Learning Theory}, Shipra Agrawal and Aaron Roth, Eds., 2024, vol. 247 of {\em Proceedings of Machine Learning Research}, pp. 1715--1754.

\bibitem{DBLP:journals/pami/RenHG017}
Shaoqing Ren, Kaiming He, Ross~B. Girshick, and Jian Sun,
\newblock ``Faster {R-CNN:} towards real-time object detection with region proposal networks,''
\newblock {\em {IEEE} Transactions on Pattern Analysis and Machine Intelligence}, vol. 39, no. 6, pp. 1137--1149, 2017.

\bibitem{DBLP:conf/cvpr/RedmonF17}
Joseph Redmon and Ali Farhadi,
\newblock ``{YOLO9000:} better, faster, stronger,''
\newblock in {\em {IEEE/CVF} Computer Vision and Pattern Recognition}, 2017, pp. 6517--6525.

\bibitem{Halmos_1974}
Paul~R Halmos,
\newblock {\em Naive set theory},
\newblock Springer-Verlag, 1974.

\bibitem{Nguyen2023SAWITAS}
Thi Thu~Thuy Nguyen, Anne~C. Eichholtzer, Don~A. Driscoll, Nathan~I. Semianiw, Dean~M. Corva, Abbas~Z. Kouzani, Thanh~Thi Nguyen, and Duc~Thanh Nguyen,
\newblock ``{SAWIT}: A small-sized animal wild image dataset with annotations,''
\newblock {\em Multimedia Tools and Applications}, vol. 83, pp. 34083--34108, 2024.

\bibitem{yolov8_ultralytics}
Glenn Jocher, Ayush Chaurasia, and Jing Qiu,
\newblock ``Ultralytics yolov8,'' 2023.

\end{thebibliography}

\end{document}